\documentclass{article}

\usepackage{microtype}
\usepackage{graphicx}
\usepackage{subcaption}
\usepackage{booktabs} 

\usepackage{booktabs}
\usepackage{multirow}
\usepackage{makecell}

\usepackage[preprint, nohyperref]{icml2026}

\usepackage{amsmath}
\usepackage{amssymb}
\usepackage{mathtools}
\usepackage{amsthm}

\theoremstyle{plain}

\theoremstyle{definition}

\theoremstyle{remark}

\usepackage[textsize=tiny]{todonotes}

\usepackage{xcolor} 
\definecolor{mycolor}{RGB}{34, 139, 34}
\usepackage{hyperref}
\hypersetup{
  colorlinks=true,
  linkcolor=red,
  urlcolor=blue,
  citecolor=mycolor
}
\usepackage[capitalize,noabbrev]{cleveref}
\begin{document}

\twocolumn[
  \icmltitle{FreeText: Training-Free Text Rendering in Diffusion Transformers via Attention Localization and Spectral Glyph Injection}



  \icmlsetsymbol{equal}{*}

  \begin{icmlauthorlist}
    \icmlauthor{Ruiqiang Zhang}{equal,yyy}
    \icmlauthor{Hengyi Wang}{equal,yyy}
    \icmlauthor{Chang Liu}{yyy}
    \icmlauthor{Guanjie Wang}{yyy}
    \icmlauthor{Zehua Ma}{yyy}
    \icmlauthor{Weiming Zhang}{yyy}
  \end{icmlauthorlist}

  \icmlaffiliation{yyy}{Anhui Province Key Laboratory of Digital Security, University of Science and Technology of China}

  \icmlcorrespondingauthor{Zehua Ma}{mzh045@ustc.edu.cn}
  \icmlcorrespondingauthor{Weiming Zhang}{zhangwm@ustc.edu.cn}

  \icmlkeywords{Machine Learning, ICML}

  \vskip 0.3in
]



\printAffiliationsAndNotice{}  

\begin{abstract}
Large-scale text-to-image (T2I) diffusion models excel at open-domain synthesis but still struggle with precise text rendering, especially for multi-line layouts, dense typography, and long-tailed scripts such as Chinese. Prior solutions typically require costly retraining or rigid external layout constraints, which can degrade aesthetics and limit flexibility. We propose \textbf{FreeText}, a training-free, plug-and-play framework that improves text rendering by exploiting intrinsic mechanisms of \emph{Diffusion Transformer (DiT)} models. \textbf{FreeText} decomposes the problem into \emph{where to write} and \emph{what to write}. For \emph{where to write}, we localize writing regions by reading token-wise spatial attribution from endogenous image-to-text attention, using sink-like tokens as stable spatial anchors and topology-aware refinement to produce high-confidence masks. For \emph{what to write}, we introduce Spectral-Modulated Glyph Injection (SGMI), which injects a noise-aligned glyph prior with frequency-domain band-pass modulation to strengthen glyph structure and suppress semantic leakage (rendering the concept instead of the word). Extensive experiments on Qwen-Image, FLUX.1-dev, and SD3 variants across longText-Benchmark, CVTG, and our CLT-Bench show consistent gains in text readability while largely preserving semantic alignment and aesthetic quality, with modest inference overhead.
\end{abstract}

\section{Introduction}
In recent years, large-scale text-to-image (T2I) diffusion models (e.g., Stable Diffusion \cite{esser2024scaling}, FLUX \cite{labs2025flux}, and Qwen-Image \cite{wu2025qwen}) have achieved strong open-domain image synthesis quality. However, precise text rendering remains challenging, with typos, missing strokes, distortions, and ``semantic drift'' (rendering the concept instead of the word), especially in multi-line, text-dense, multilingual, and semantically complex scenes. The issue is particularly severe for logographic scripts such as Chinese: the character distribution is highly long-tailed with many rare characters and low-frequency compositions underrepresented during training; meanwhile, numerous characters are visually similar with complex internal radicals and stroke patterns \cite{chen2021zero}. As a result, models often fail to learn reliable glyph priors from limited coverage and are prone to fine-grained confusion, making the rendered text frequently unusable even after repeated sampling.

From both application and research perspectives, text rendering is not a cosmetic add-on but a key stress test for fine-grained controllability, complex scene planning, and cross-modal alignment in T2I models. Text is a highly structured visual object whose strokes, glyph shapes, and arrangements impose strict local geometry and global layout constraints. Moreover, humans are extremely sensitive to textual errors: in real-world scenarios such as posters and UI design, text often serves as a crucial identifier, and typos or malformed glyphs can severely degrade usability. Therefore, better text rendering is essential for practical usability, where minor typos can invalidate an otherwise good image.

Most existing approaches to improve text rendering rely on two ingredients: additional training or fine-tuning (retraining-based) and explicit layout or position conditions (layout-conditioned). Methods such as TextDiffuser \cite{chen2023textdiffuser} and AnyText \cite{tuo2023anytext} train layout predictors or control branches with box/mask/glyph supervision, improving controllability and OCR accuracy. These methods incur high data/compute costs and often shift the generation distribution and visual style away from the base model. At inference time, they further inject bounding boxes, masks, or glyphs as hard conditions, mechanically fixing text regions to preset positions. Such external constraints can suppress the model's intrinsic scene-planning behavior, making it difficult to balance diversity and naturalness under complex backgrounds or ambiguous/conflicting prompts.

\begin{figure}[t]
  \centering
  \includegraphics[width=\linewidth]{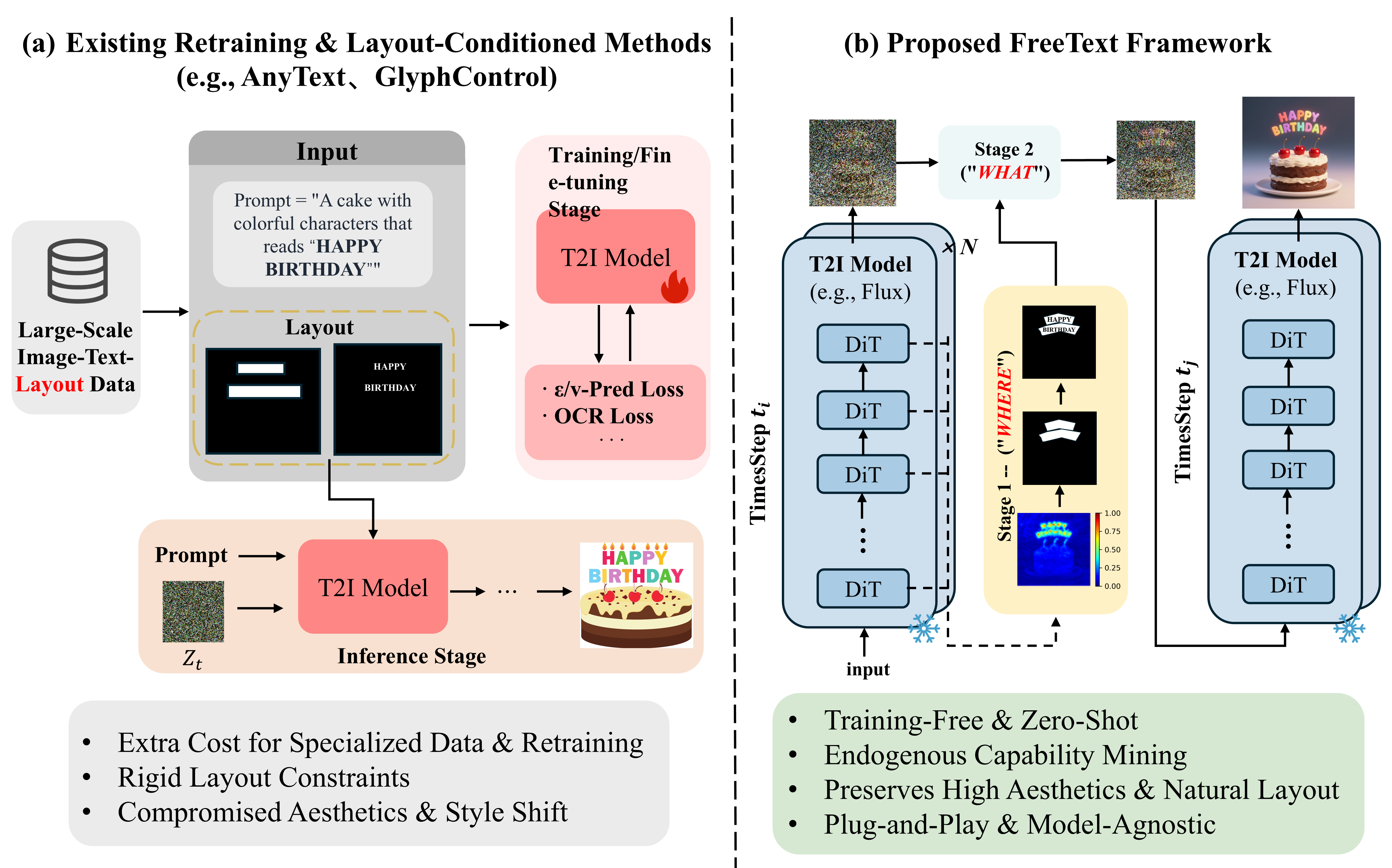}
  \caption{System overview. (a) Prior text-rendering methods typically require retraining and/or rigid layout conditions. (b) FreeText decomposes text rendering into \emph{WHERE} and \emph{WHAT}: it localizes text regions via endogenous attention maps, then injects a glyph-structure prior in a model-compatible way, enabling training-free enhancement while preserving the base model's aesthetics.}
  \label{fig:system_overview}
\end{figure}

Meanwhile, large fully pre-trained and extensively post-trained models (e.g., FLUX and Qwen-Image) already exhibit strong aesthetic quality; imposing rigid layout constraints on them is not only difficult to fine-tune but can also noticeably damage their aesthetics. Conversely, text-specialized models such as AnyText, which are trained primarily for text rendering, typically cannot replicate the full pre-training and post-training pipelines of large foundation models, and thus often trade rendering accuracy against aesthetics, with the latter lagging in complex open-domain scenes. To date, there remains limited progress on simultaneously achieving high rendering accuracy and strong aesthetics by leveraging only the base model's internal mechanisms, without modifying architectures or parameters.

Motivated by these limitations, we propose a new perspective: instead of paying the high cost of retraining to teach models a generic \textit{how to write}, we decompose text rendering into two more fundamental subproblems that the base model already has the potential to support: \textit{where to write} and \textit{what to write}. Based on this view, we introduce FreeText, a training-free, plug-and-play enhancement framework. The design is driven by a key observation: it is much easier for the model to recognize text than to precisely render text (pixel-level glyph generation). FreeText exploits easily accessible visual priors of text and the model's internal structure to address these two subproblems.

\begin{enumerate}
  \item \emph{WHERE}\textit{ to write.} T2I models are not necessarily lacking layout planning; rather, we have not effectively read out their internal plans for text regions. In fact, during generation, diffusion models with DiT-style architectures implicitly encode spatial attribution for different text tokens in image-to-text cross-attention \cite{peebles2023scalable}. Attention maps across timesteps and network depths jointly describe the model's endogenous layout. Based on this, we propose an unsupervised localization strategy: instead of relying on fragile external OCR or Vision-Language Model(VLM) detectors for post-hoc detection, FreeText selects the most stable attention layers as spatial anchors and precisely locks the writing regions for target text tokens under zero layout annotations (zero-layout supervision).

  \item \emph{WHAT}\textit{ to write.} As illustrated in Appendix Fig.~1, models may render the token ``Car'' as a car image rather than the word itself. We attribute this to the coupling between semantic concepts (high-level meaning) and glyph structures (visual form) in the embedding space. Early in generation, strong concrete semantic priors can dominate and suppress glyph information, causing semantic leakage---i.e., concepts overwhelm strokes and lead to ``text becoming images''. To enforce the local rule ``Glyph $>$ Semantics'', we propose Spectral-Modulated Glyph Injection (SGMI). Instead of naively mixing latents, SGMI applies band-pass modulation in the frequency domain to enhance mid-to-high frequency components that carry glyph structures, while suppressing the propagation of background and irrelevant noise, thereby guiding accurate glyph synthesis.
\end{enumerate}

In summary, our contributions are:
\begin{itemize}
  \item A training-free, base-model-agnostic text rendering enhancement framework. FreeText operates as an inference-time plug-in, seamlessly integrating into Stable Diffusion, FLUX, Qwen-Image, and other T2I models without modifying any parameters, and substantially improves text rendering performance in bilingual (Chinese/English) and challenging rendering scenarios.
  \item An unsupervised text-region localization method based on endogenous attention. We leverage DiT-style image-to-text attention signals and an Attention Sink-like stability cue to achieve generic and high-precision text-region locking without any supervision.
  \item A frequency-domain glyph prior injection scheme. SGMI uses band-pass spectral modulation to emphasize structure-carrying glyph frequencies while suppressing semantic-background leakage, improving rendering fidelity.
  \item A Chinese long-tail text rendering benchmark. We introduce CLT-Bench, a graded evaluation benchmark targeting long-tail Chinese characters (rare and structurally complex) to systematically assess performance degradation from common to rare, and from simple to complex settings.
\end{itemize}

\section{Related Work}

\subsection{T2I diffusion foundation models}
Recent large-scale T2I diffusion models have steadily improved resolution, semantic alignment, and text rendering \cite{wu2025qwen, seedream2025seedream, esser2024scaling, labs2025flux}. Representative systems such as Stable Diffusion 3, Qwen-Image, and FLUX.1 attribute these gains to stronger MMDiT/DiT backbones, flow/rectified-flow objectives, and large dedicated data pipelines, resulting in better overall visual quality and typography. However, such improvements typically require costly pre-training and post-training, and are tightly coupled to specific architectures and data recipes, making text-rendering capability hard to transfer across base models at low cost. In contrast, FreeText keeps the base model unchanged and performs inference-time control by leveraging endogenous attention and latent-space structure, enabling cross-model, fine-grained text rendering enhancement.

\subsection{Retraining and layout-dependent text rendering}
Most prior text-rendering methods follow a retraining-based, layout-dependent paradigm. TextDiffuser-style \cite{chen2023textdiffuser} approaches learn layout prediction modules on large OCR-annotated corpora, requiring explicit layout templates or segmentation priors at generation time. Methods such as AnyText \cite{tuo2023anytext}, GlyphDraw \cite{ma2023glyphdraw}, GlyphControl \cite{yang2023glyphcontrol}, and UniGlyph \cite{yang2023glyphcontrol} introduce ControlNet-style or dedicated conditional branches on top of Stable Diffusion/DiT, retraining with extra inputs (e.g., glyph images, text masks, or segmentation maps) to improve OCR accuracy and font controllability. While effective, these methods rely on additional annotations and control branches, tightly binding generation to external layout/visual conditions, limiting prompt freedom and image diversity, and underutilizing the base model's endogenous scene planning. FreeText is training-free and layout-free: it localizes text regions from endogenous attention and injects glyph priors via spectral modulation, improving text rendering without any additional training cost.

\subsection{Attention sinks}
Attention has long served as a lens for interpreting Transformer behavior. In large language models, the attention sink phenomenon has been widely observed: a few semantically weak tokens absorb disproportionate attention, stabilizing inference by buffering global context \cite{tigges2023linear, razzhigaev2025llm, chauhan2025punctuation, zhangattention}. Related analyses in multimodal models use attention patterns to study cross-modal alignment and hallucination \cite{kang2025see}. Yet, attention sinks have rarely been exploited for spatial generation, and have not been systematically used for text-region localization in T2I diffusion models. FreeText empirically finds that sink-like tokens in DiT-based T2I models produce relatively stable boundary cues across timesteps and layers, and treats them as spatial anchors to extract text regions from endogenous image-to-text attention without supervision, providing reliable localization for subsequent glyph prior injection.

\section{Method}
\begin{figure*}[t]
  \centering
  \includegraphics[width=\textwidth]{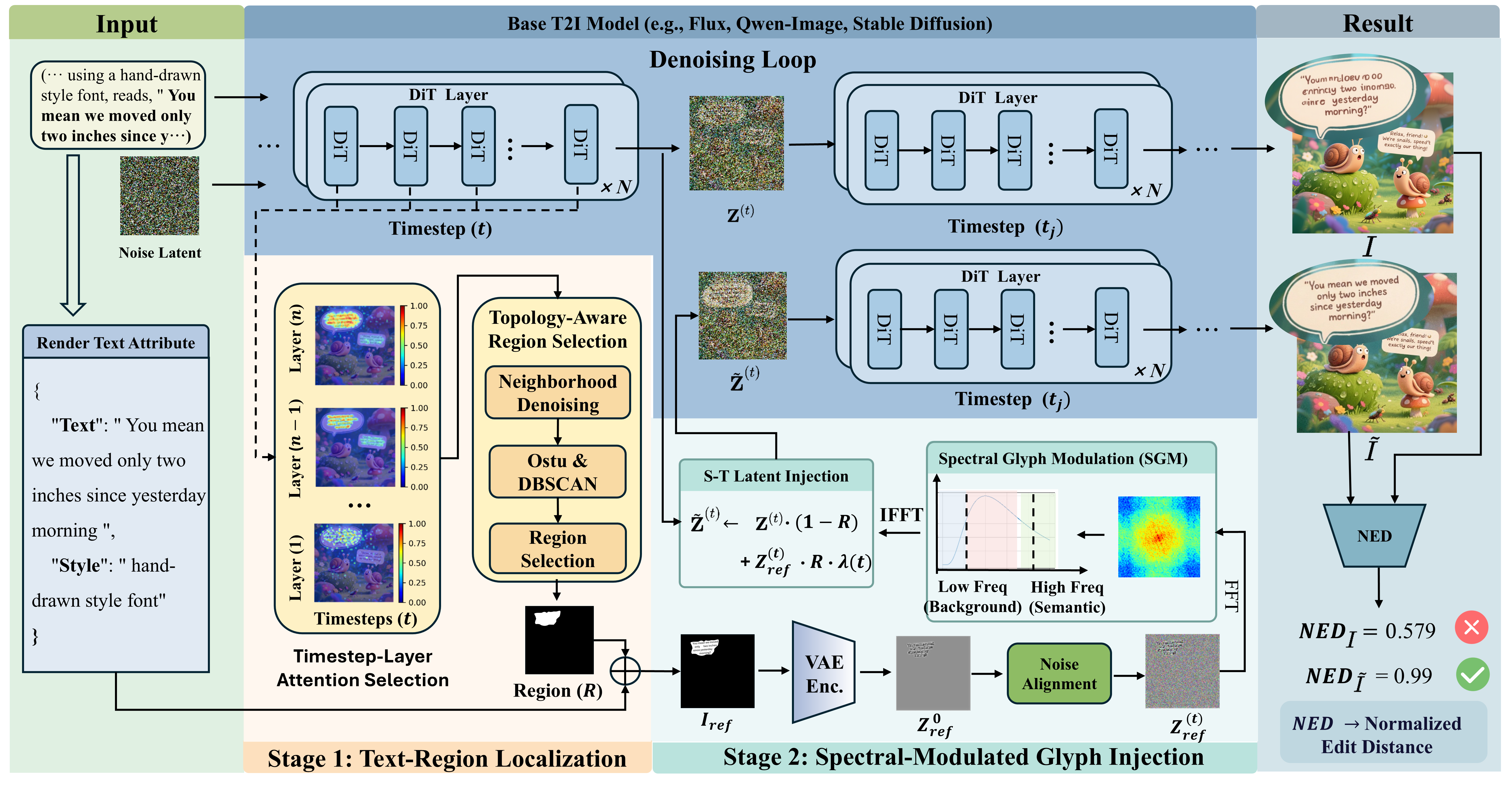}
  \caption{Overview of FreeText.}
  \label{method_overview}
\end{figure*}

\label{sec:method}

FreeText aims to enhance text rendering in complex scenes without modifying the architecture or parameters of a base T2I diffusion model. Given a target text span $s$ and its glyph reference image, FreeText proceeds in two stages, as shown in Fig.~\ref{method_overview}.
\begin{enumerate}
  \item \textbf{Attention-guided endogenous text-region localization} (Sec.~\ref{sec:where}): we extract image-to-text (I2T) cross-attention from DiT/MMDiT blocks during sampling, aggregate and select informative timestep--layer pairs, and apply topology-aware post-processing to obtain a high-confidence writing mask $\mathbf{R}_s$ in latent space.
  \item \textbf{Spectral-Modulated Glyph Injection} (Sec.~\ref{sec:what}): we encode the glyph reference into latent space, align it to the current noise level, construct a Log-Gabor based Spectral-Modulated Glyph Injection (SGMI) prior, and inject it into $\mathbf{R}_s$ within a short time window using cosine annealing, strengthening glyph structure and suppressing semantic leakage (e.g., rendering the concept instead of the word).
\end{enumerate}

\subsection{Attention-guided text-region localization}
\label{sec:where}
To answer ``where to write'', we localize the writing region directly from endogenous attention \cite{tang2023daam}, without external layout predictors, OCR, or VLM detectors. We read out token-wise spatial attribution from attention maps, then perform timestep--layer selection and topology-aware refinement to produce a high-confidence region mask.

\subsubsection{Attention extraction}
Let $\mathbf{A}^{(t,l)}$ denote the head-averaged I2T attention at timestep $t$ and the $l$-th DiT/MMDiT block:
\begin{equation}
\mathbf{A}^{(t,l)} \in \mathbb{R}^{H \times W \times N_{\text{text}}},
\end{equation}
where $N_{\text{text}}$ is the number of text tokens. For a target span $s$, we first locate its token subsequence $\mathcal{T}_s$, and augment it with a few sink-like special tokens that exhibit stable high responses across layers/heads. We call the union the anchor token set $\tilde{\mathcal{T}}_s$.

We then average attention over $\tilde{\mathcal{T}}_s$ to obtain an initial localization map:
\begin{equation}
\mathbf{M}^{(t,l)}(x,y)=\frac{1}{|\tilde{\mathcal{T}}_s|}\sum_{k\in \tilde{\mathcal{T}}_s}\mathbf{A}^{(t,l)}_{x,y,k},
\end{equation}
and linearly normalize $\mathbf{M}^{(t,l)}$ to $[0,1]$. For clarity, we omit the subscript $s$ in what follows.

\subsubsection{Timestep-layer selection}
\begin{figure}[ht]
  \centering
  \includegraphics[width=\linewidth]{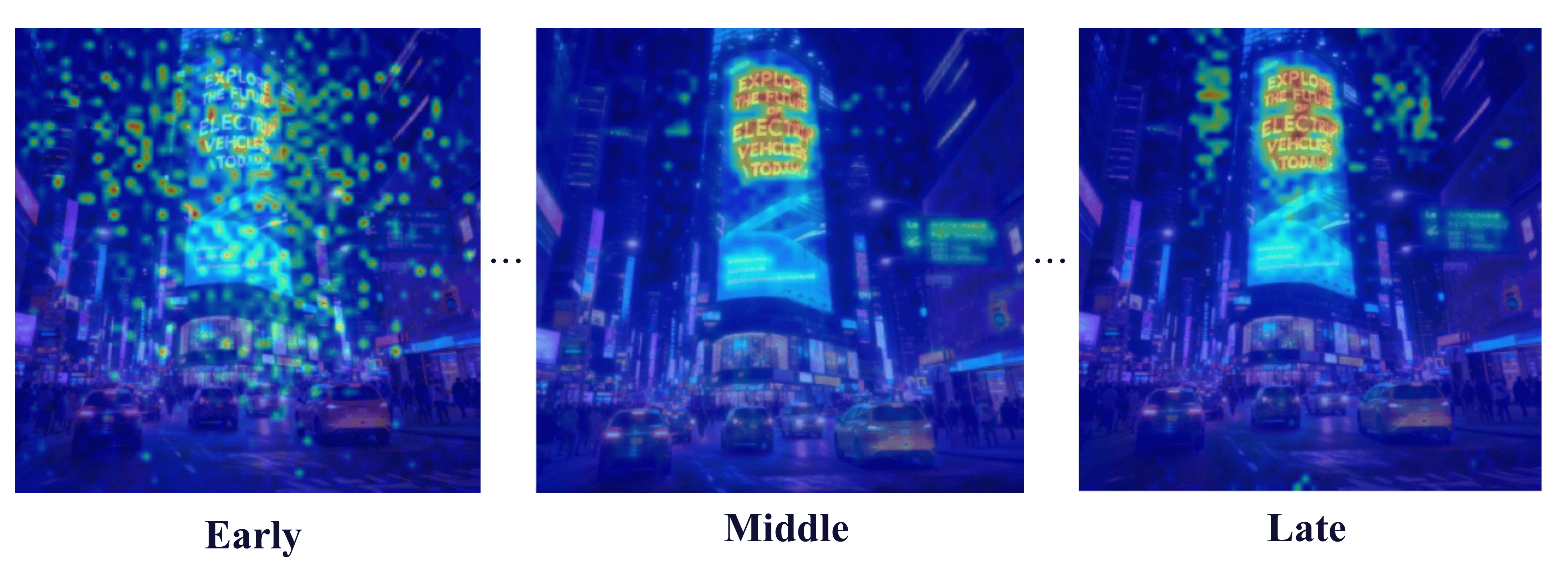}
  \caption{Typical I2T attention patterns across timesteps: early steps are coarse, mid steps concentrate on target regions, and late steps become diffuse.}
  \label{fig:attn_over_t}
\end{figure}

As shown in Fig.~\ref{fig:attn_over_t}, naively aggregating attention across all timesteps and blocks introduces substantial noise: early steps are coarse and reflect global planning; mid steps are most informative for writing placement; late steps become diffuse due to global refinement \cite{chefer2023attend, darcet2023vision}. In addition, shallow blocks emphasize local geometry while deeper blocks integrate global semantics. We therefore select informative timestep-layer pairs before aggregation.

Given candidate sets $\mathcal{T}_{\text{cand}}$ and $\mathcal{L}_{\text{cand}}$, we score each pair $(t,l)$ using a \emph{soft IoU} between $\mathbf{M}^{(t,l)}$ and a reference mask $\mathbf{Y}\in[0,1]^{H\times W}$:

\begin{equation}
\text{IoU}(t,l)=
\frac{\langle \mathbf{M}^{(t,l)},\mathbf{Y}\rangle}
{\|\mathbf{M}^{(t,l)}\|_{1}+\|\mathbf{Y}\|_{1}-\langle \mathbf{M}^{(t,l)},\mathbf{Y}\rangle}.
\end{equation}

We select the top-$K$ pairs to form $\mathcal{S}$ and aggregate:
\begin{equation}
\mathbf{M}(x,y)=\frac{1}{|\mathcal{S}|}\sum_{(t,l)\in\mathcal{S}}\mathbf{M}^{(t,l)}(x,y).
\end{equation}

\subsubsection{Topology-aware region selection}
The aggregated map $\mathbf{M}$ may still contain isolated peaks and fragmented clusters. We apply a lightweight post-processing pipeline to produce the final writing mask.

We first perform local neighborhood aggregation on $\mathbf{M}$ to suppress small outliers and promote connected responses. Next, we binarize $\mathbf{M}$ into $\mathbf{B}\in\{0,1\}^{H\times W}$ using an adaptive threshold selected by maximizing inter-class variance \cite{otsu1975threshold}. We then run DBSCAN \cite{ester1996density} on foreground pixels to obtain candidate connected regions $\{\mathcal{C}_i\}$ while discarding sparse noise.

Each region $\mathcal{C}_i$ is scored on the original $\mathbf{M}$:
\begin{equation}
q_i=\frac{\left|\{(x,y)\in \mathcal{C}_i \mid \mathbf{M}(x,y)>\tau\}\right|}{|\mathcal{C}_i|},
\end{equation}
where $\tau$ is set as a high quantile of $\mathbf{M}$ within the union of candidate regions. We select the best region and resize it to latent resolution to obtain the binary writing mask:
\begin{equation}
\mathbf{R}\in \{0,1\}^{H_{\text{lat}}\times W_{\text{lat}}}.
\end{equation}
In Sec.~\ref{sec:what}, $\mathbf{R}$ is broadcast across channels for local latent injection.

\subsection{Spectral-Modulated Glyph Injection}
\label{sec:what}
To answer ``what to write'', we enhance glyph structure while suppressing semantic leakage. We encode a glyph reference into latent space, align it to the current noise level, apply Log-Gabor based SGMI to emphasize structure-carrying frequencies, and inject the resulting prior into $\mathbf{R}$ within a short time window.

\subsubsection{Noise-aligned latent projection}
We rasterize the target text $s$ into a glyph reference image $\mathbf{I}_{\text{glyph}}$ placed in region $\mathbf{R}$, and encode it with the same VAE as the base model:
\begin{equation}
\mathbf{z}_{\text{ref}}=E_{\text{VAE}}(\mathbf{I}_{\text{glyph}})
\in \mathbb{R}^{C\times H_{\text{lat}}\times W_{\text{lat}}}.
\end{equation}
At timestep $t$ with noise schedule $(\alpha_t,\sigma_t)$, we match the noise level via forward diffusion:
\begin{equation}
\mathbf{z}_{\text{ref}}^{(t)}=\alpha_t\,\mathbf{z}_{\text{ref}}+\sigma_t\,\boldsymbol{\epsilon},
\quad \boldsymbol{\epsilon}\sim\mathcal{N}(0,\mathbf{I}).
\end{equation}

\subsubsection{Log-Gabor spectral modulation}
On $\mathbf{z}_{\text{ref}}^{(t)}$, we apply a Log-Gabor filter \cite{field1987relations} to strengthen mid-to-high frequencies that carry glyph structure while suppressing low-frequency background and ultra-high-frequency noise. Let $G(\rho,\theta)$ be the Log-Gabor kernel in the 2D frequency domain. For each channel $c$:
\begin{align}
\widehat{\mathbf{z}}_{\text{ref},c}^{(t)} &= \mathcal{F}\!\left(\mathbf{z}_{\text{ref},c}^{(t)}\right), \\
\widehat{\mathbf{z}}_{\text{sgmi},c}^{(t)}(\rho,\theta) &= G(\rho,\theta)\,\cdot \widehat{\mathbf{z}}_{\text{ref},c}^{(t)}(\rho,\theta), \\
\mathbf{z}_{\text{sgmi},c}^{(t)} &= \mathcal{F}^{-1}\!\left(\widehat{\mathbf{z}}_{\text{sgmi},c}^{(t)}\right),
\end{align}
where $\mathcal{F}$ and $\mathcal{F}^{-1}$ are 2D FFT and inverse FFT. The resulting $\mathbf{z}_{\text{sgmi}}^{(t)}$ is the SGMI-enhanced reference latent at timestep $t$.

\subsubsection{Annealed spatiotemporal injection}
Let the sampling trajectory evolve from timestep $T$ to $0$. We inject glyph priors only in a mid-early window:
\begin{equation}
t_{\text{start}}=0.8T,\quad t_{\text{end}}=0.6T,
\end{equation}
to avoid disrupting early global planning or late-stage fine-detail refinement. For $t\in[t_{\text{start}},t_{\text{end}}]$, we define a cosine-annealed weight:
\begin{equation}
\lambda(t)=\frac{1}{2}\left(1+\cos\left(\pi\cdot\frac{t-t_{\text{start}}}{t_{\text{end}}-t_{\text{start}}}\right)\right),
\end{equation}
and update the denoising latent $\mathbf{z}^{(t)}$ by masked replacement \cite{avrahami2023blended}:

\begin{equation}
\tilde{\mathbf{z}}^{(t)} =
\big(\mathbf{I}-\lambda(t)\mathbf{R}\big)\odot \mathbf{z}^{(t)} \;+\;
\lambda(t)\mathbf{R}\odot \mathbf{z}_{\text{sgmi}}^{(t)} .
\end{equation}

For $t\notin[t_{\text{start}},t_{\text{end}}]$, we keep $\mathbf{z}^{(t)}$ unchanged.

\subsection{CLT-Bench: Chinese long-tail text rendering}
\label{sec:cltbench}

Chinese text rendering is challenging due to a long-tailed character distribution and high intra-class visual similarity. Existing benchmarks over-emphasize common characters and/or English, obscuring degradation from frequent/simple to rare/complex cases \cite{zhao2025lex, fang2025flux, du2025textcrafter}. We introduce CLT-Bench to stress-test T2I text rendering under rare-character and complex-layout settings.

We assign each prompt a complexity score combining character difficulty and layout difficulty. For a character $c$, we normalize stroke count $s(c)$ and frequency rank $r(c)$:
For a character $c$, we normalize stroke count $\kappa(c)$ and frequency rank $r(c)$:
\begin{equation}
K(c)=\frac{\kappa(c)-\kappa_{\min}}{\kappa_{\max}-\kappa_{\min}},\quad
R(c)=\frac{r(c)-r_{\min}}{r_{\max}-r_{\min}}.
\end{equation}
and define character difficulty
\begin{equation}
D(c)=\frac{w_s K(c)+w_f R(c)}{w_s+w_f}\in[0,1].
\end{equation}
Given text segments $\{\mathrm{txt}_i\}_{i=1}^{N_{\text{seg}}}$ with characters $\{c_j\}_{j=1}^{N_{\text{chars}}}$, we compute
\begin{equation}
\begin{aligned}
C_{\text{char}}&=\frac{1}{N_{\text{chars}}}\sum_{j}D(c_j),\\
C_{\text{len}}&=\min\!\left(\frac{N_{\text{chars}}}{N_{\max}},1\right),\\
C_{\text{seg}}&=\min\!\left(\frac{N_{\text{seg}}-1}{M_{\max}-1},1\right),
\end{aligned}
\end{equation}
where $N_{\max}$ is a preset upper bound on the total number of characters to render in a prompt, and $M_{\max}$ is a preset upper bound on the number of text segments (regions) to render. The prompt score is then
\begin{equation}
\emph{Score}=\frac{w_{\text{char}}C_{\text{char}}+w_{\text{len}}C_{\text{len}}+w_{\text{seg}}C_{\text{seg}}}{w_{\text{char}}+w_{\text{len}}+w_{\text{seg}}}\in[0,1].
\end{equation}
We stratify prompts by \emph{Score} to form subsets spanning common/simple to rare/complex characters with challenging multi-segment layouts.

\begin{figure*}[t]
  \centering
  \includegraphics[width=\textwidth]{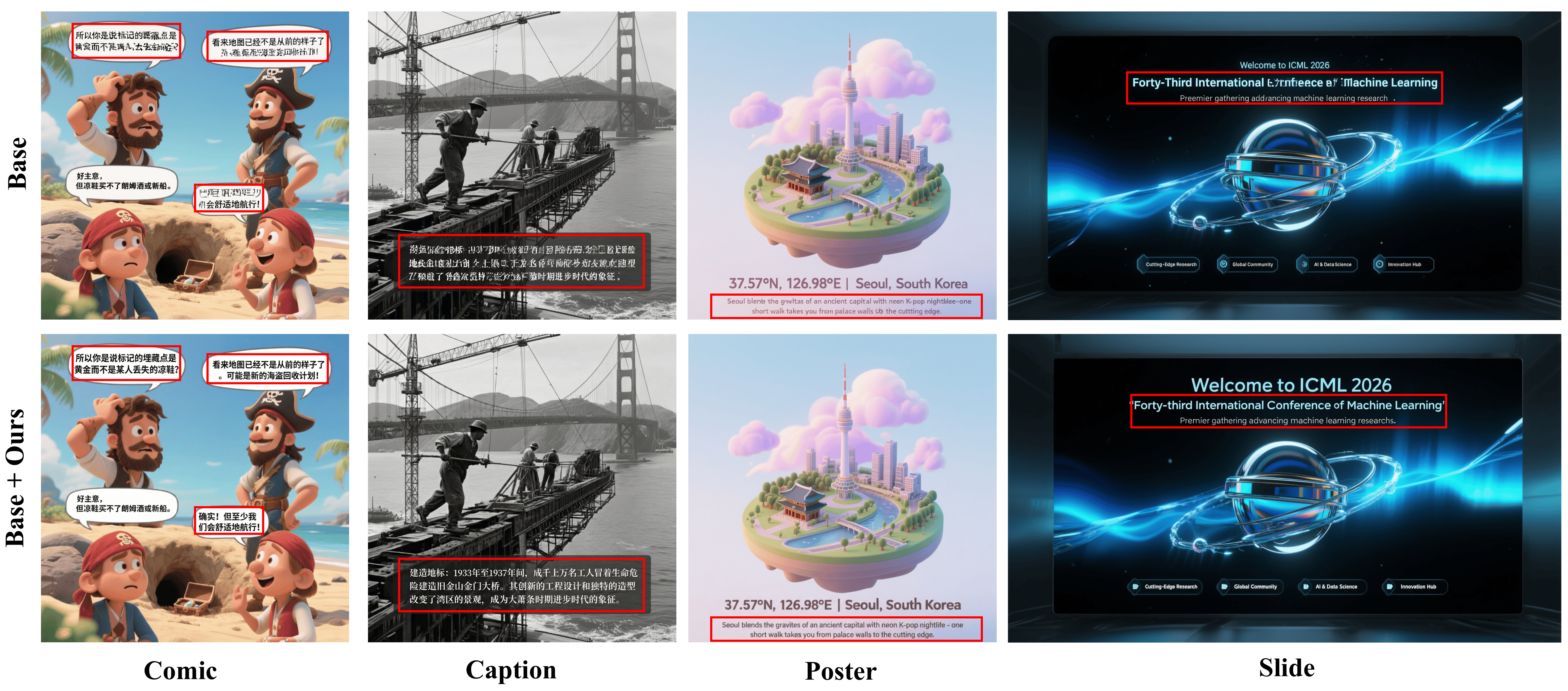}
  \caption{Baseline comparison across four text-rendering scenarios (comic, caption, poster, slide). Top: Base; bottom: Base+FreeText. Red boxes highlight the target text regions, where FreeText reduces typos/malformed glyphs and improves readability.}
  \label{fig:baseline_compare}
\end{figure*}

\newpage
\section{Experiments}
\label{sec:experiments}

\subsection{Experimental setup}
\label{sec:exp_setup}

\subsubsection{Base models}
We evaluate FreeText on four representative T2I foundation models:
(i) Qwen-Image (Chinese/English prompts),
(ii) FLUX.1-dev (English only),
(iii) Stable Diffusion 3.5 Large (SD3.5-L; English only),
and (iv) Stable Diffusion 3 Medium (SD3-M; English only).
All experiments compare \emph{Base} vs.\ \emph{Base + FreeText}. FreeText is used as an inference-time plug-in: it does not modify model parameters, architectures, or introduce learnable branches.

\subsubsection{Benchmarks and protocol}
We use three benchmarks covering long text, multi-region rendering, and long-tail Chinese:
(1) \textbf{longText-Benchmark} with longText-en/zh, focusing on long prompts and paragraph-level, multi-line text \cite{geng2025x}; 
(2) \textbf{CVTG}, with 2/3/4/5 text regions (2--5 segments) and typically short prompts \cite{du2025textcrafter};
(3) \textbf{CLT-Bench} (Sec.~\ref{sec:cltbench}), targeting rare and structurally complex Chinese characters.

\textbf{Language alignment.}
Qwen-Image and FLUX.1-dev are evaluated on longText-Benchmark and CVTG.
SD3.5-L and SD3-M are evaluated on CVTG only, since long prompts can be truncated by their text encoders.
CLT-Bench is evaluated on Qwen-Image only (Chinese support).

\textbf{Inference settings.}
Unless noted, Base and Base + FreeText use identical resolution, sampling steps, and sampler hyperparameters.
FreeText uses the default annealed injection window (Sec.~\ref{sec:what}); in this section we refer to the injection module as \textbf{SGMI}.

\subsubsection{Metrics}
We measure both text readability and overall image quality (higher is better unless noted):
\textbf{NED} (Normalized Edit Distance, via a fixed OCR engine \cite{cui2025paddleocr}),
\textbf{CLIPScore} (text--image alignment),
\textbf{AestheticScore} (LAION aesthetic predictor),
and \textbf{VQA Score} (VLM-based usability/clarity QA; templates in the appendix).
For localization analysis, we report \textbf{IoU} between predicted and reference text regions.

\subsection{Effectiveness of FreeText}
\label{sec:exp_effectiveness}

\subsubsection{Qwen-Image and FLUX.1-dev}
Table~\ref{tab:longtext} reports results on longText-Benchmark and CVTG. FreeText consistently improves NED and VQA Score \cite{fang2025flux}, indicating higher text readability, while CLIPScore and AestheticScore remain largely stable, suggesting limited impact on semantic alignment and aesthetics.

\begin{table*}[ht]
  \centering
  \small
  \setlength{\tabcolsep}{5.5pt}
  \renewcommand{\arraystretch}{1.1}
  \caption{End-to-end results on longText-Benchmark and CVTG.}
  \label{tab:longtext}
  \begin{tabular}{l l l c c c c}
    \toprule
    Model & Setting & Subset & NED$\uparrow$ & CLIP$\uparrow$ & Aes$\uparrow$ & VQA$\uparrow$ \\
    \midrule
    Qwen-Image & Base & longText-en & 0.625 & 0.858 & 4.912 & 2.650 \\
    Qwen-Image & Base + FreeText & longText-en & \textbf{0.713} & \textbf{0.864} & \textbf{5.013} & \textbf{4.177} \\
    \addlinespace[2pt]
    FLUX.1-dev & Base & longText-en & 0.598 & 0.863 & \textbf{5.365} & 2.563 \\
    FLUX.1-dev & Base + FreeText & longText-en & \textbf{0.690} & \textbf{0.868} & 5.342 & \textbf{4.211} \\
    \midrule
    Qwen-Image & Base & longText-zh & 0.639 & 0.474 & 4.607 & 3.657 \\
    Qwen-Image & Base + FreeText & longText-zh & \textbf{0.694} & \textbf{0.537} & \textbf{4.749} & \textbf{4.211} \\
    \midrule
    Qwen-Image & Base & CVTG & 0.574 & 0.781 & 4.386 & 2.756 \\
    Qwen-Image & Base + FreeText & CVTG & \textbf{0.619} & \textbf{0.794} & \textbf{4.391} & \textbf{3.469} \\
    \addlinespace[2pt]
    FLUX.1-dev & Base & CVTG & 0.712 & 0.836 & 5.910 & 4.050 \\
    FLUX.1-dev & Base + FreeText & CVTG & \textbf{0.722} & \textbf{0.839} & \textbf{5.936} & \textbf{4.952} \\
    \bottomrule
  \end{tabular}
\end{table*}

\subsubsection{SD3-M and SD3.5-L}
Since SD3 variants are sensitive to long prompts, we evaluate them on CVTG only (Table~\ref{tab:sd3_cvtg}). FreeText improves NED and VQA Score for both models, while CLIPScore and AestheticScore remain comparable, indicating the local SGMI injection does not introduce notable semantic drift or quality degradation.

\begin{table}[ht]
  \centering
  \small
  \setlength{\tabcolsep}{4.2pt}
  \renewcommand{\arraystretch}{1.1}
  \caption{End-to-end results on CVTG for SD3 models. Best within each model pair is in bold.}
  \label{tab:sd3_cvtg}
  \begin{tabular}{l l c c c c}
    \toprule
    Model & Setting & NED$\uparrow$ & CLIP$\uparrow$ & Aes$\uparrow$ & VQA$\uparrow$ \\
    \midrule
    SD3.5-L & Base & 0.848 & \textbf{0.879} & \textbf{5.634} & 3.849 \\
    SD3.5-L & Base + FreeText & \textbf{0.864} & 0.871 & 5.608 & \textbf{4.595} \\
    \midrule
    SD3-M & Base & 0.616 & 0.851 & 5.906 & 2.903 \\
    SD3-M & Base + FreeText & \textbf{0.669} & \textbf{0.852} & \textbf{5.917} & \textbf{3.674} \\
    \bottomrule
  \end{tabular}
\end{table}

\subsubsection{CLT-Bench}
On CLT-Bench (Qwen-Image only), FreeText improves NED but with smaller gains (Table~\ref{tab:clt}). This suggests SGMI is most effective when the base model already has a usable representation for the target characters; it strengthens glyph structure rather than enabling unseen characters from scratch.

\subsubsection{Benefit propagation under full attention}
We observe cross-region benefit propagation: correcting one text region with FreeText can improve other regions that are not explicitly processed, reflected by higher global metrics (e.g., VQA Score). We attribute this to global self-attention in DiT/MMDiT: patch tokens mix information globally at each denoising step, so severe errors in one region can perturb updates elsewhere; once a key error is corrected, this interference is reduced.

\begin{table}[ht]
  \centering
  \small
  \setlength{\tabcolsep}{6pt}
  \renewcommand{\arraystretch}{1.1}
  \caption{End-to-end NED on CLT-Bench.}
  \label{tab:clt}
  \begin{tabular}{l l c}
    \toprule
    Model & Setting & NED$\uparrow$ \\
    \midrule
    Qwen-Image & Base & 0.458 \\
    Qwen-Image & Base + FreeText & \textbf{0.488} \\
    \bottomrule
  \end{tabular}
\end{table}

\begin{figure}[ht]
  \centering
  \includegraphics[width=0.9\columnwidth]{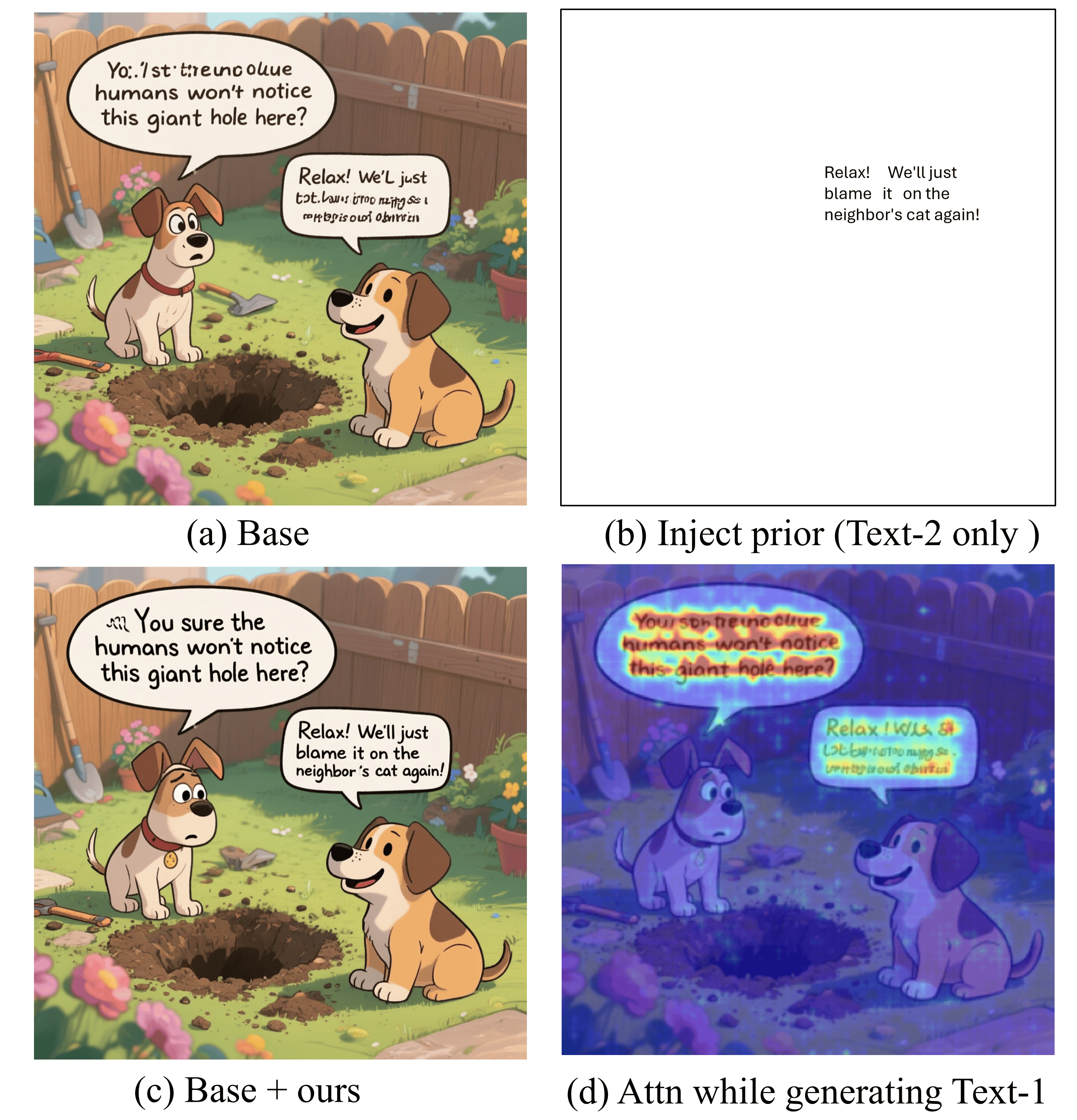}
  \caption{Cross-region benefit propagation and attention evidence (example with two text lines; refining only one line can improve the other).}
  \label{fig:benefit_propagation}
\end{figure}

\subsection{Localization strategy}
\label{sec:exp_localization}

\subsubsection{Token choice}
\begin{figure}[ht]
  \centering
  \includegraphics[width=\columnwidth]{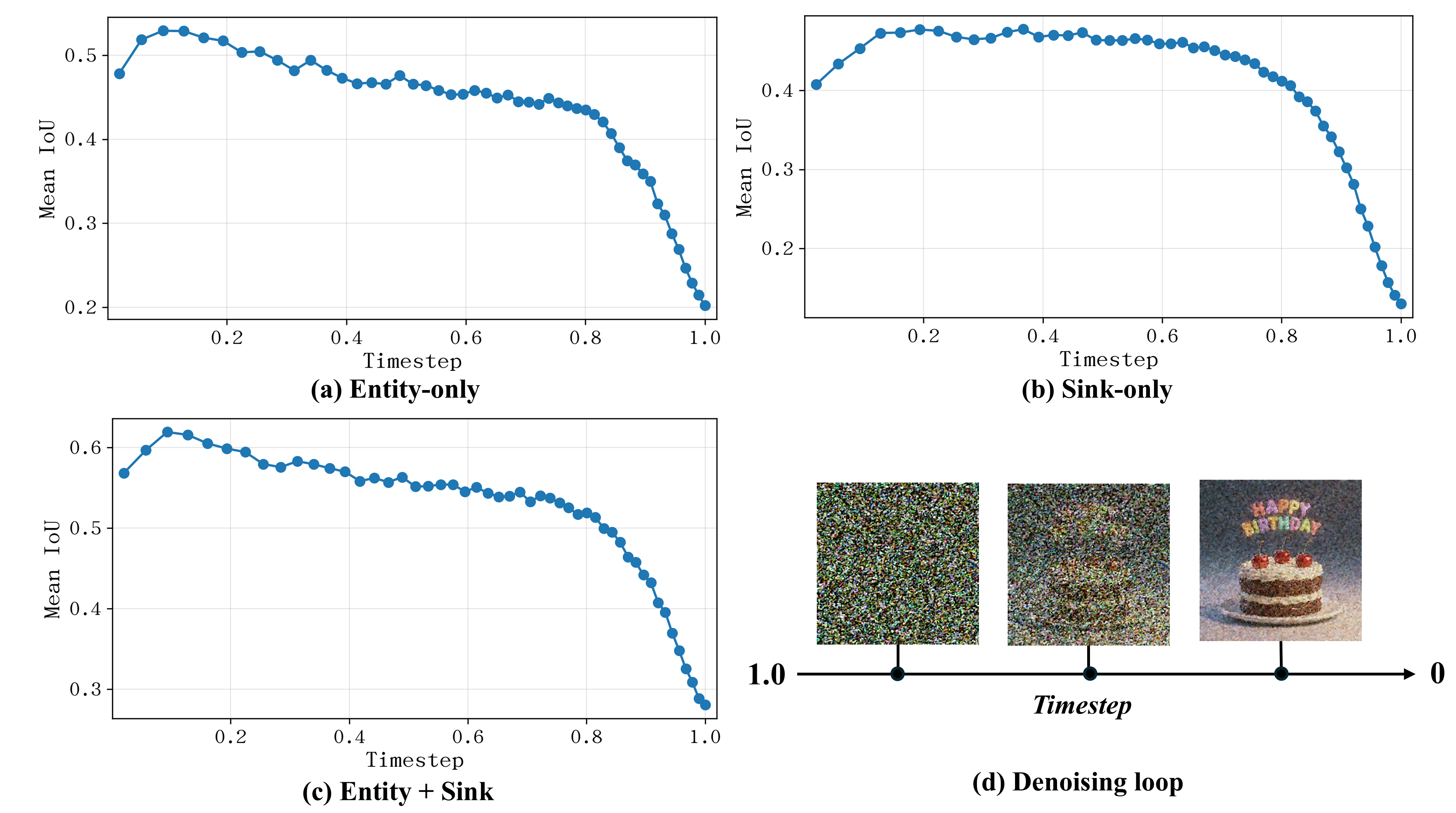}
  \caption{IoU vs.\ timestep for different token sets.}
  \label{fig:token_curve}
\end{figure}

We compare three token sets for each target span: \textbf{Entity-only} (tokens of the target string), \textbf{Sink-only} (sink-like special tokens), and \textbf{Entity + Sink}. As shown in Table~\ref{tab:token_choice} and Fig.~\ref{fig:token_curve}, Sink-only is more temporally stable but has a lower ceiling, while Entity + Sink achieves the best IoU by combining explicit semantic attribution with stable sink responses, yielding more reliable masks for SGMI.

\begin{table}[ht]
  \centering
  \small
  \setlength{\tabcolsep}{20pt}
  \renewcommand{\arraystretch}{1.1}
  \caption{Localization IoU for different token sets.}
  \label{tab:token_choice}
  \begin{tabular}{l c}
    \toprule
    Setting & IoU$\uparrow$ \\
    \midrule
    Entity-only & 0.495 \\
    Sink-only & 0.479 \\
    Entity + Sink & \textbf{0.561} \\
    \bottomrule
  \end{tabular}
\end{table}

\subsubsection{Comparison with VLM-based localization}
Table~\ref{tab:vlm_loc} compares our endogenous localization against several closed-source VLM baselines. In practice, multi-line text, cluttered backgrounds, and malformed glyphs can break ``recognize-then-localize'' pipelines; recognition failure often cascades into localization failure. By reading I2T attention directly, FreeText avoids this chain and provides a more stable signal.

\begin{table}[ht]
  \centering
  \small
  \setlength{\tabcolsep}{18pt}
  \renewcommand{\arraystretch}{1.1}
  \caption{Localization IoU comparison.}
  \label{tab:vlm_loc}
  \begin{tabular}{l c}
    \toprule
    Method & IoU$\uparrow$ \\
    \midrule
    doubao-seed-1-6-251015 & 0.325 \\
    gemini-2.5-flash-lite & 0.139 \\
    gpt-5.1 & 0.159 \\
    qwen3-vl-plus-2025-09-23 & 0.195 \\
    FreeText (ours) & \textbf{0.561} \\
    \bottomrule
  \end{tabular}
\end{table}

\begin{figure}[ht]
  \centering
  \includegraphics[width=\columnwidth]{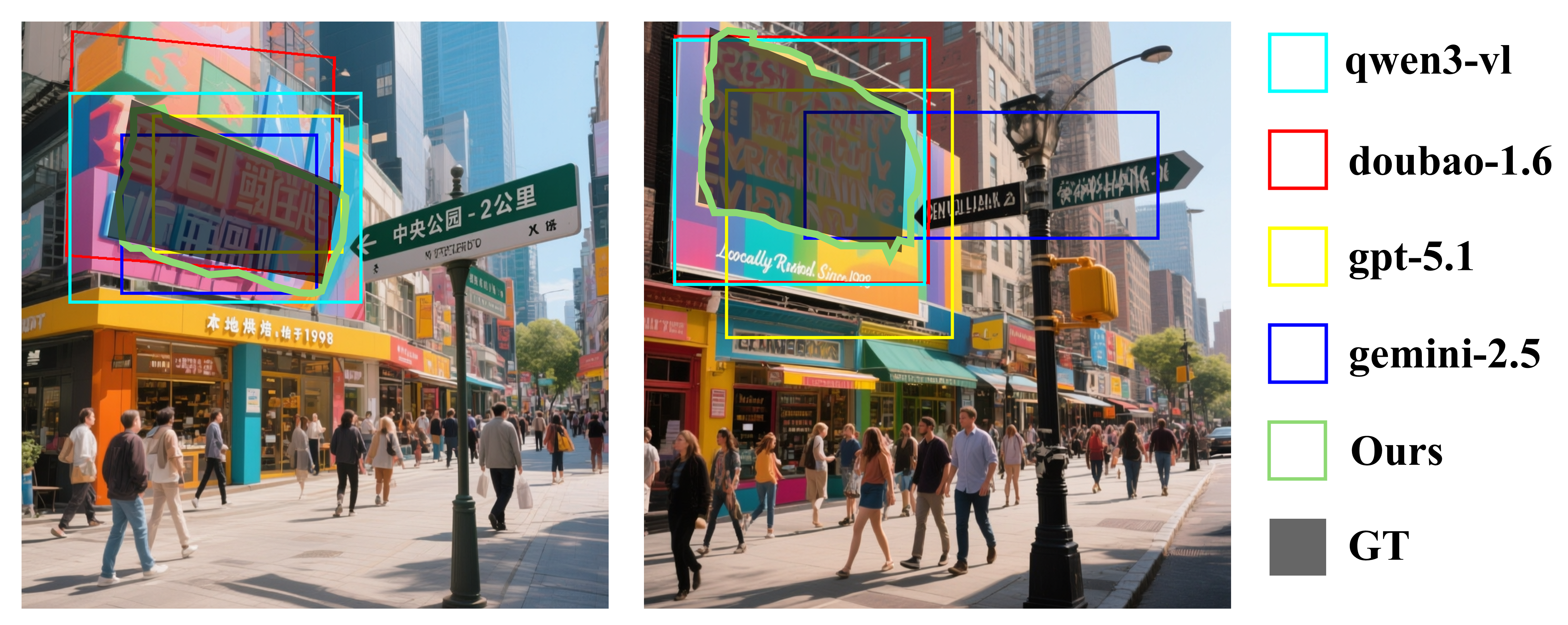}
  \caption{Typical VLM localization failures under multi-line text and degraded glyphs, compared with endogenous localization.}
  \label{fig:vlm_fail}
\end{figure}

\subsection{Ablation study}
\label{sec:exp_ablation}
\begin{figure}[ht]
  \centering
  \includegraphics[width=0.9\columnwidth]{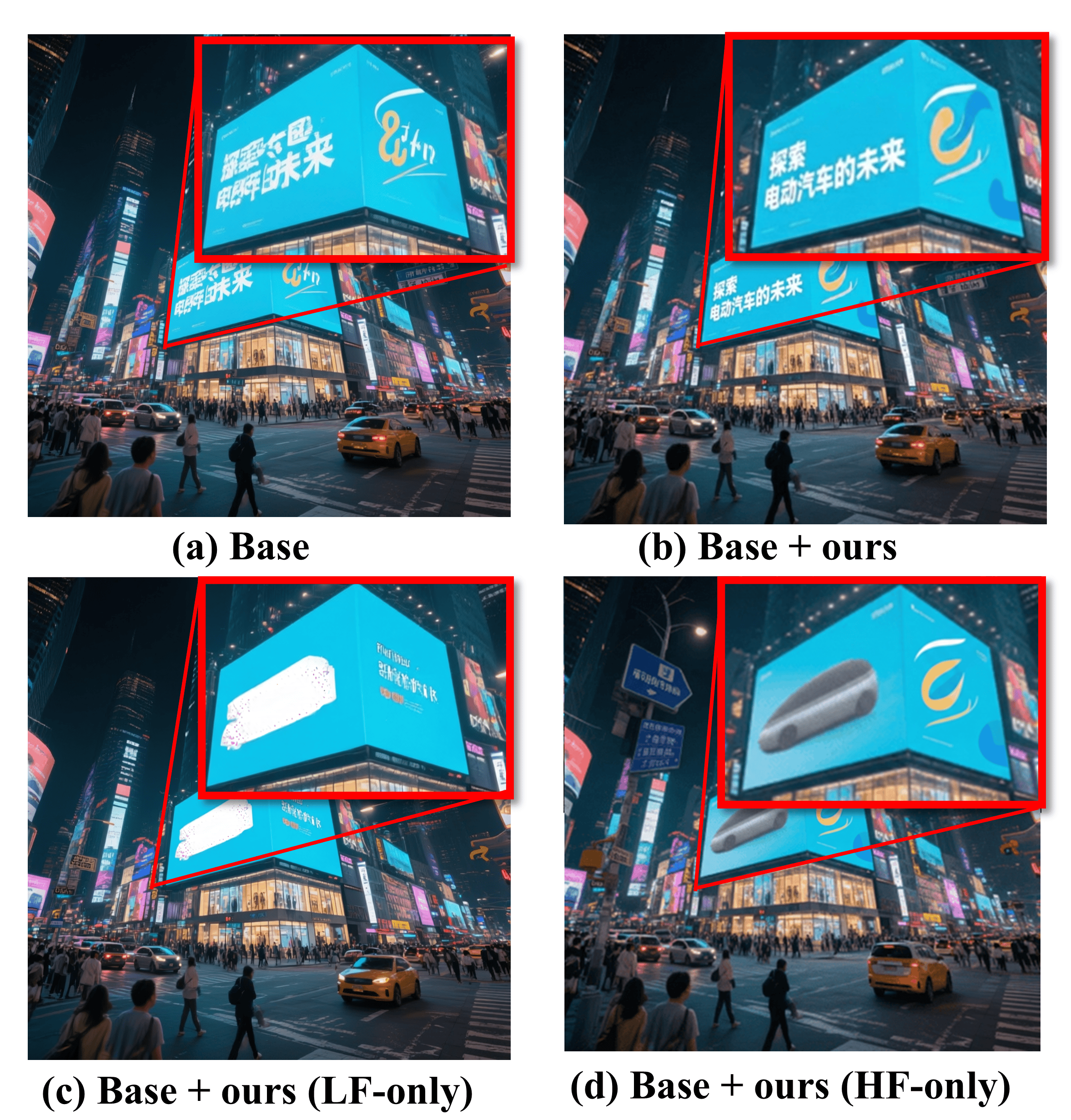}
  \caption{Qualitative ablation illustrating semantic leakage and stroke degradation under different spectral settings.}
  \label{fig:freq_ablation}
\end{figure}

We compare three variants: \textbf{B} (Base), \textbf{+F} (Base+FreeText), and \textbf{+F$-\,$SGMI} (Base+FreeText without SGMI, i.e., removing the spectral band-pass modulation while keeping the rest unchanged). As shown in Table~\ref{tab:ablation_sgmi}, removing SGMI reduces NED and VQA Score, while CLIP and Aes remain largely unchanged, indicating SGMI primarily contributes to text readability improvements. As further illustrated in Fig.~\ref{fig:freq_ablation}, injecting only low-frequency components loses stroke-level structure, while injecting only high-frequency components (where semantics dominates) can trigger concept-texture intrusion. In contrast, SGMI's band-pass design provides an injection signal that is most effective for glyph structure while being most conservative against semantic leakage. This indicates that the key of frequency-domain modulation is not \emph{injecting more information}, but \emph{injecting the right spectral band}.

\begin{table}[ht]
  \centering
  \small
  \setlength{\tabcolsep}{4.2pt}
  \renewcommand{\arraystretch}{1.08}
  \caption{Ablation on SGMI.}
  \label{tab:ablation_sgmi}
  \begin{tabular}{l c c c c c}
    \toprule
    Model & Settings & NED$\uparrow$ & CLIP$\uparrow$ & Aes$\uparrow$ & VQA$\uparrow$ \\
    \midrule
    \multirow{3}{*}{Qwen-Image} 
      & B & 0.625 & 0.858 & 4.912 & 2.650 \\
      & +F$-\,$SGMI & 0.686 & 0.860 & \textbf{5.027} & 3.724 \\
      & +F & \textbf{0.713} & \textbf{0.864} & 5.013 & \textbf{4.177} \\
    \midrule
    \multirow{3}{*}{FLUX.1-dev} 
      & B & 0.598 & 0.863 & \textbf{5.365} & 2.563 \\
      & +F$-\,$SGMI & 0.671 & 0.865 & 5.361 & 3.816 \\
      & +F & \textbf{0.690} & \textbf{0.868} & 5.342 & \textbf{4.211} \\
    \bottomrule
  \end{tabular}
\end{table}

\begin{table}[h]
  \centering
  \small
  \setlength{\tabcolsep}{4.6pt}
  \renewcommand{\arraystretch}{1.1}
  \caption{Inference efficiency.}
  \label{tab:efficiency}
  \begin{tabular}{l l c c}
    \toprule
    Model & Setting & Time (s)$\downarrow$ & Mem (GB)$\downarrow$ \\
    \midrule
    Qwen-Image & Base & \textbf{37.64} & \textbf{53.76} \\
    Qwen-Image & Base + FreeText & 42.33 & 54.35 \\
    \midrule
    FLUX.1-dev & Base & \textbf{41.56} & \textbf{31.44} \\
    FLUX.1-dev & Base + FreeText & 47.17 & 32.17 \\
    \midrule
    SD3.5-L & Base & \textbf{35.03} & \textbf{26.11} \\
    SD3.5-L & Base + FreeText & 41.17 & 26.91 \\
    \midrule
    SD3-M & Base & \textbf{9.85} & \textbf{14.53} \\
    SD3-M & Base + FreeText & 11.47 & 14.97 \\
    \bottomrule
  \end{tabular}
\end{table}
\subsection{Efficiency}
\label{sec:exp_efficiency}
We measure inference overhead on an NVIDIA A6000 with bfloat16, resolution $928\times928$, and 50 sampling steps. Table~\ref{tab:efficiency} shows that FreeText adds moderate overhead (primarily from Stage-1 localization, which accumulates and selects I2T attention before injection), increasing end-to-end latency by roughly 12\%--18\% with $<1$GB peak-memory overhead.

\section{Conclusion}
\label{sec:conclusion}

We presented FreeText, a training-free and base-model-agnostic framework for improving text rendering in T2I diffusion models without changing model weights or architectures. By decomposing text rendering into \emph{where to write} and \emph{what to write}, FreeText (i) localizes writing regions from endogenous attention via sink-anchored, topology-aware selection, and (ii) enhances glyph fidelity through SGMI, a noise-aligned frequency-domain injection that strengthens structure-carrying components and mitigates semantic leakage. Across multiple foundation models and benchmarks, FreeText consistently improves readability metrics while maintaining CLIPScore and AestheticScore, and incurs only moderate runtime and memory overhead. Future research will focus on validating the universality of our approach by adapting it to diverse emerging foundation models.

\bibliography{example_paper}

@article{wu2025qwen,
  title={Qwen-image technical report},
  author={Wu, Chenfei and Li, Jiahao and Zhou, Jingren and Lin, Junyang and Gao, Kaiyuan and Yan, Kun and Yin, Sheng-ming and Bai, Shuai and Xu, Xiao and Chen, Yilei and others},
  journal={arXiv preprint arXiv:2508.02324},
  year={2025}
}

@article{seedream2025seedream,
  title={Seedream 4.0: Toward next-generation multimodal image generation},
  author={Seedream, Team and Chen, Yunpeng and Gao, Yu and Gong, Lixue and Guo, Meng and Guo, Qiushan and Guo, Zhiyao and Hou, Xiaoxia and Huang, Weilin and Huang, Yixuan and others},
  journal={arXiv preprint arXiv:2509.20427},
  year={2025}
}

@inproceedings{esser2024scaling,
  title={Scaling rectified flow transformers for high-resolution image synthesis},
  author={Esser, Patrick and Kulal, Sumith and Blattmann, Andreas and Entezari, Rahim and M{\"u}ller, Jonas and Saini, Harry and Levi, Yam and Lorenz, Dominik and Sauer, Axel and Boesel, Frederic and others},
  booktitle={Forty-first international conference on machine learning},
  year={2024}
}

@article{labs2025flux,
  title={FLUX. 1 Kontext: Flow Matching for In-Context Image Generation and Editing in Latent Space},
  author={Labs, Black Forest and Batifol, Stephen and Blattmann, Andreas and Boesel, Frederic and Consul, Saksham and Diagne, Cyril and Dockhorn, Tim and English, Jack and English, Zion and Esser, Patrick and others},
  journal={arXiv preprint arXiv:2506.15742},
  year={2025}
}

@article{tuo2023anytext,
  title={Anytext: Multilingual visual text generation and editing},
  author={Tuo, Yuxiang and Xiang, Wangmeng and He, Jun-Yan and Geng, Yifeng and Xie, Xuansong},
  journal={arXiv preprint arXiv:2311.03054},
  year={2023}
}

@article{chen2023textdiffuser,
  title={Textdiffuser: Diffusion models as text painters},
  author={Chen, Jingye and Huang, Yupan and Lv, Tengchao and Cui, Lei and Chen, Qifeng and Wei, Furu},
  journal={Advances in Neural Information Processing Systems},
  volume={36},
  pages={9353--9387},
  year={2023}
}

@article{chen2021zero,
  title={Zero-shot Chinese character recognition with stroke-level decomposition},
  author={Chen, Jingye and Li, Bin and Xue, Xiangyang},
  journal={arXiv preprint arXiv:2106.11613},
  year={2021}
}

@article{yang2023glyphcontrol,
  title={Glyphcontrol: Glyph conditional control for visual text generation},
  author={Yang, Yukang and Gui, Dongnan and Yuan, Yuhui and Liang, Weicong and Ding, Haisong and Hu, Han and Chen, Kai},
  journal={Advances in Neural Information Processing Systems},
  volume={36},
  pages={44050--44066},
  year={2023}
}

@article{ma2023glyphdraw,
  title={Glyphdraw: Seamlessly rendering text with intricate spatial structures in text-to-image generation},
  author={Ma, Jian and Zhao, Mingjun and Chen, Chen and Wang, Ruichen and Niu, Di and Lu, Haonan and Lin, Xiaodong},
  journal={arXiv preprint arXiv:2303.17870},
  year={2023}
}

@article{tigges2023linear,
  title={Linear representations of sentiment in large language models, 2023},
  author={Tigges, Curt and Hollinsworth, Oskar John and Geiger, Atticus and Nanda, Neel},
  journal={URL https://arxiv. org/abs/2310.15154},
  year={2023}
}

@article{razzhigaev2025llm,
  title={Llm-microscope: Uncovering the hidden role of punctuation in context memory of transformers},
  author={Razzhigaev, Anton and Mikhalchuk, Matvey and Rahmatullaev, Temurbek and Goncharova, Elizaveta and Druzhinina, Polina and Oseledets, Ivan and Kuznetsov, Andrey},
  journal={arXiv preprint arXiv:2502.15007},
  year={2025}
}

@article{chauhan2025punctuation,
  title={Punctuation and predicates in language models},
  author={Chauhan, Sonakshi and Chaudhary, Maheep and Choy, Koby and Nellessen, Samuel and Schoots, Nandi},
  journal={arXiv preprint arXiv:2508.14067},
  year={2025}
}

@inproceedings{zhangattention,
  title={Attention Sinks: A'Catch, Tag, Release'Mechanism for Embeddings},
  author={Zhang, Stephen and Khan, Mustafa and Papyan, Vardan},
  booktitle={The Thirty-ninth Annual Conference on Neural Information Processing Systems},
  year={2025}
}

@article{kang2025see,
  title={See what you are told: Visual attention sink in large multimodal models},
  author={Kang, Seil and Kim, Jinyeong and Kim, Junhyeok and Hwang, Seong Jae},
  journal={arXiv preprint arXiv:2503.03321},
  year={2025}
}

@inproceedings{ester1996density,
  title={A density-based algorithm for discovering clusters in large spatial databases with noise},
  author={Ester, Martin and Kriegel, Hans-Peter and Sander, J{\"o}rg and Xu, Xiaowei and others},
  booktitle={kdd},
  volume={96},
  number={34},
  pages={226--231},
  year={1996}
}

@article{otsu1975threshold,
  title={A threshold selection method from gray-level histograms},
  author={Otsu, Nobuyuki and others},
  journal={Automatica},
  volume={11},
  number={285-296},
  pages={23--27},
  year={1975}
}

@article{zhao2025lex,
  title={LeX-Art: Rethinking Text Generation via Scalable High-Quality Data Synthesis},
  author={Zhao, Shitian and Wu, Qilong and Li, Xinyue and Zhang, Bo and Li, Ming and Qin, Qi and Liu, Dongyang and Zhang, Kaipeng and Li, Hongsheng and Qiao, Yu and others},
  journal={arXiv preprint arXiv:2503.21749},
  year={2025}
}

@article{fang2025flux,
  title={Flux-reason-6m \& prism-bench: A million-scale text-to-image reasoning dataset and comprehensive benchmark},
  author={Fang, Rongyao and Yu, Aldrich and Duan, Chengqi and Huang, Linjiang and Bai, Shuai and Cai, Yuxuan and Wang, Kun and Liu, Si and Liu, Xihui and Li, Hongsheng},
  journal={arXiv preprint arXiv:2509.09680},
  year={2025}
}

@article{du2025textcrafter,
  title={Textcrafter: Accurately rendering multiple texts in complex visual scenes},
  author={Du, Nikai and Chen, Zhennan and Gao, Shan and Chen, Zhizhou and Chen, Xi and Jiang, Zhengkai and Yang, Jian and Tai, Ying},
  journal={arXiv preprint arXiv:2503.23461},
  year={2025}
}

@article{geng2025x,
  title={X-omni: Reinforcement learning makes discrete autoregressive image generative models great again},
  author={Geng, Zigang and Wang, Yibing and Ma, Yeyao and Li, Chen and Rao, Yongming and Gu, Shuyang and Zhong, Zhao and Lu, Qinglin and Hu, Han and Zhang, Xiaosong and others},
  journal={arXiv preprint arXiv:2507.22058},
  year={2025}
}

@article{cui2025paddleocr,
  title={Paddleocr 3.0 technical report},
  author={Cui, Cheng and Sun, Ting and Lin, Manhui and Gao, Tingquan and Zhang, Yubo and Liu, Jiaxuan and Wang, Xueqing and Zhang, Zelun and Zhou, Changda and Liu, Hongen and others},
  journal={arXiv preprint arXiv:2507.05595},
  year={2025}
}

@article{field1987relations,
  title={Relations between the statistics of natural images and the response properties of cortical cells},
  author={Field, David J},
  journal={Journal of the Optical Society of America A},
  volume={4},
  number={12},
  pages={2379--2394},
  year={1987},
  publisher={OSA}
}

@article{chefer2023attend,
  title={Attend-and-excite: Attention-based semantic guidance for text-to-image diffusion models},
  author={Chefer, Hila and Alaluf, Yuval and Vinker, Yael and Wolf, Lior and Cohen-Or, Daniel},
  journal={ACM transactions on Graphics (TOG)},
  volume={42},
  number={4},
  pages={1--10},
  year={2023},
  publisher={ACM New York, NY, USA}
}

@inproceedings{tang2023daam,
  title={What the daam: Interpreting stable diffusion using cross attention},
  author={Tang, Raphael and Liu, Linqing and Pandey, Akshat and Jiang, Zhiying and Yang, Gefei and Kumar, Karun and Stenetorp, Pontus and Lin, Jimmy and T{\"u}re, Ferhan},
  booktitle={Proceedings of the 61st Annual Meeting of the Association for Computational Linguistics (Volume 1: Long Papers)},
  pages={5644--5659},
  year={2023}
}

@article{darcet2023vision,
  title={Vision transformers need registers},
  author={Darcet, Timoth{\'e}e and Oquab, Maxime and Mairal, Julien and Bojanowski, Piotr},
  journal={arXiv preprint arXiv:2309.16588},
  year={2023}
}

@article{avrahami2023blended,
  title={Blended latent diffusion},
  author={Avrahami, Omri and Fried, Ohad and Lischinski, Dani},
  journal={ACM transactions on graphics (TOG)},
  volume={42},
  number={4},
  pages={1--11},
  year={2023},
  publisher={ACM New York, NY, USA}
}

@inproceedings{peebles2023scalable,
  title={Scalable diffusion models with transformers},
  author={Peebles, William and Xie, Saining},
  booktitle={Proceedings of the IEEE/CVF international conference on computer vision},
  pages={4195--4205},
  year={2023}
}
\bibliographystyle{icml2026}

\newpage
\appendix
\onecolumn



\end{document}